# From Questions to Queries: An AI-powered Multi-Agent Framework for Spatial Text-to-SQL

Ali Khosravi Kazazi[a], Zhenlong Li[a*], M. Naser Lessani[a], Guido Cervone[b]

[a] Geoinformation and Big Data Research Laboratory, Department of Geography, The Pennsylvania State University, University Park, PA, USA

[b] Institute for Computational and Data Sciences and Department of Geography, The Pennsylvania State University, University Park, PA, USA

[*]Correspondence: zhenlong@psu.edu

**Abstract**

The complexity of SQL and the spatial semantics of PostGIS create barriers for non-experts working with spatial data. Although large language models can translate natural language into SQL, spatial Text-to-SQL is more error-prone than general Text-to-SQL because it must resolve geographic intent, schema ambiguity, geometry-bearing tables and columns, spatial function choice, and coordinate reference system and measurement assumptions. We introduce a multi-agent framework that addresses these coupled challenges through staged interpretation, schema grounding, logical planning, SQL generation, and execution-based review. The framework is supported by a knowledge base with programmatic schema profiling, semantic enrichment, and embedding-based retrieval. We evaluated the framework on the non-spatial KaggleDBQA benchmark and on SpatialQueryQA, a new multi-level and coverage-oriented benchmark with diverse geometry types, workload categories, and spatial operations. On KaggleDBQA, the system reached 81.2% accuracy, 221 of 272 questions, after reviewer corrections. On SpatialQueryQA, the system achieved 87.7% accuracy, 79 of 90, compared with 76.7% without the review stage. These results show that decomposing the task into specialized but tightly coupled agents improves robustness, especially for spatially sensitive queries. The study improves access to spatial analysis and provides a practical step toward more reliable spatial Text-to-SQL systems and Autonomous GIS.

## 1. Introduction

The ability to collect data has grown significantly across various domains, with more than half of it being geospatially referenced (Hahmann & Burghardt, 2013). Geospatial data is paramount as organizations and businesses seek to leverage it for critical decision-making (Erskine et al., 2013). Areas like urban planning, health and transportation are strongly dependent on geospatial data analysis and visualization (Joshi et al., 2012; Loidl et al., 2016; Rinner, 2007). Robust geospatial data analysis relies on precise data retrieval that integrates multi-source datasets, mitigates information overload, and facilities semantics-aware interpretation, thereby enabling intelligent decision-making (J. Liu et al., 2021).

Today, a wide range of relational database management systems are designed or adopted for storing, manipulating, retrieving or even analyzing geospatial data. Notable examples include Google BigQuery GIS (Mozumder & Karthikeya, 2022), PostGIS (Obe & Hsu, 2011), Oracle with the Spatial and Graph options (Sankaranarayanan et al., 2009) and Amazon S3. Among these, the first three systems rely on Structured Query Language (SQL) for querying and data management. In addition, SQL remains one of the most widely used languages within GIScience community (Ramezan et al., 2024). However, non-experts often find SQL overwhelming and susceptible to mistakes in geospatial data management. Therefore, experts who possess both SQL proficiency and geoprocessing expertise are required to obtain meaningful results. Such professionals must also have a thorough understanding of the existing database schemas and table structures (Kanburoğlu & Tek, 2024).

Among the tools that support geospatial data management, PostGIS, an open-source extension for PostgreSQL, distinguishes itself with specialized geospatial operations. The extension supports spatial data types, functions, and indexing capabilities to manage geospatial data efficiently. Due to its robustness, scalability, Open Geospatial Consortium (OGC) standards compliance, and open-source nature, PostGIS has become one of the most widely adopted tools for storing and analyzing geospatial information. However, as a SQL-based system, PostGIS poses challenges for non-expert users, who must understand both SQL queries and geometry types, coordinate reference systems, spatial functions, and broader spatial literacy.

Over the last few years, generative Artificial Intelligence, particularly Large Language Models (LLMs), has shown strong ability in natural language understanding, reasoning, and code generation. Generating SQL from a natural language question is therefore one of the most important applied uses of LLMs, especially when database complexity makes manual exploration slow or impractical. However, spatial Text-to-SQL is not only a SQL generation problem. It is also a coupled problem of interpreting geographic language, grounding concepts to geometry-bearing tables and columns, selecting appropriate PostGIS operations, handling CRS- and unit-sensitive measurements, and validating whether the final query is spatially meaningful. Because these requirements often need to be satisfied at the same time, a single-pass approach is more likely to produce syntactically valid but semantically incorrect spatial SQL. Figure 1 shows the visual comparison between a single-agent workflow and a multi-agent workflow in the context of spatial SQL.

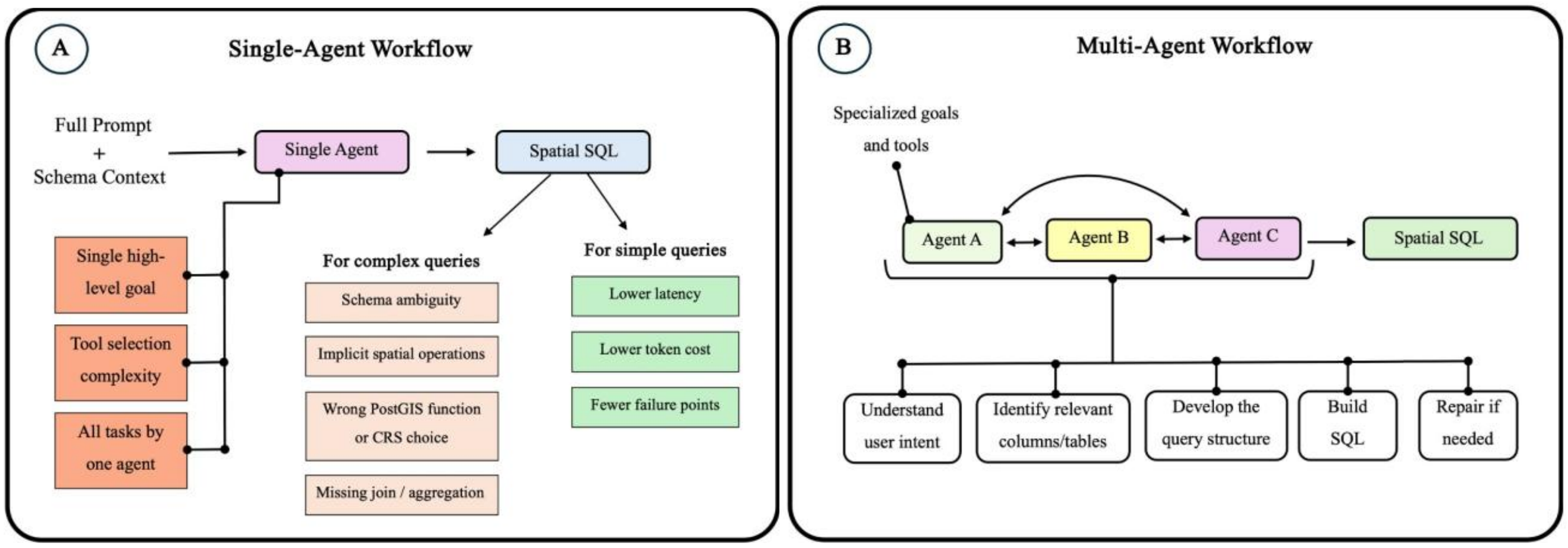

Figure 1. Visual difference between the single-agent and multi-agent systems. (A) The single-agent design handles all spatial operations and query generation at one prompt. (B) The multi-agent pipeline separates understanding, planning, and execution review into different components.

A staged multi-agent design is especially useful for harder spatial cases, although simpler queries may still be handled adequately by a strong single-agent approach. This motivation gives rise to three research questions: 1) how can a multi-agent LLM framework be designed to accurately translate natural language questions into spatial SQL queries? 2) to what extent does employing a multi-agent approach within a comprehensive system improve the robustness of spatial Text-to-SQL compared with single-agent or prompt-based approaches? and 3) what constitutes an effective multi-level and coverage-oriented benchmark for evaluating Spatial Text-to-SQL systems?

To address these questions, we propose a system centered on a tightly coupled multi-agent pipeline. The design follows four ideas. First, the task is separated by cognitive function, so understanding the question, grounding it to schema, planning the logic, generating SQL, and reviewing execution are handled by different agents. Second, grounding happens before generation, so later stages work on validated schema rather than guessed tables or columns. Third, each stage passes a structured intermediate representation to the next one, which reduces stage drift and keeps later reasoning aligned with earlier constraints. Fourth, the final SQL is checked against database behavior rather than trusted after one pass. Beyond the framework itself, the study also contributes SpatialQueryQA, a multi-level and coverage-oriented benchmark designed to evaluate spatial Text-to-SQL across diverse geometry types, workload categories, and levels of difficulty.

The remainder of the paper is organized as follows. Section 2 reviews related work on geospatial SQL queries, natural-language interfaces, and LLM-based geospatial agents. Section 3 explains the proposed methodology. Section 4 presents the evaluation design and benchmark datasets. Section 5 reports the experimental setup and results. Section 6 discusses limitations, trade-offs, and future extensions, and Section 7 concludes with implications for future geospatial Text-to-SQL research.

## 2. Related Work

The emergence of LLMs and their adoption by the GIScience community are changing the field by enabling natural language interaction, spatial analysis automation, and geospatial data retrieval (Akinboyewa et al., 2025; Li & Ning, 2023a; Wang et al., 2024). Recent work has started to use LLMs not only for query translation, but also for broader geospatial workflow automation. Research on Autonomous GIS and GIS Copilots treats the LLM as a reasoning core that can plan steps, call tools, and support non-experts inside GIS environments. These studies provide important context for the present work because they show that geospatial tasks often benefit from explicit tool use, memory, and staged reasoning rather than a single free-form prompt.

Within spatial natural-language interfaces, several directions can be identified. Prompt-based methods enrich the prompt with schema descriptions, examples, or retrieved context to guide SQL generation (Jiang & Yang, 2024). Retrieval-augmented and schema-mapping methods such as Spatial-RAG and try to improve grounding by connecting the question to relevant schema and geographic context before SQL is formed (C. Yu et al., 2025; D. Yu et al., 2025). NALSpatial frames the problem around practical query families, including range, nearest-neighbor, spatial join, and aggregation queries, and provides an important baseline for understanding common workload patterns in natural-language access to spatial databases (M. Liu et al., 2025). These studies are closely related to our goal, but they generally focus more on direct translation or retrieval design than on a full multi-agent workflow with explicit planning and execution-based review.

The literature review highlights three main gaps. First, many existing studies emphasize a narrow subset of geometry types, spatial operations, or interaction settings, so their findings do not always transfer to more heterogeneous spatial database workloads. Second, although prompt engineering, retrieval augmentation, and dynamic schema mapping have improved performance, the question of how to decompose spatial Text-to-SQL into tightly coordinated stages remains underexplored. Third, benchmark limitations remain important. Existing datasets often focus on one source, one query language, templated questions, or narrow spatial behavior. For this reason, the field still needs public benchmarks that are better coverage-oriented, with diverse geometry types, multiple workload families, and varying reasoning difficulty, even if they are not necessarily large in sample count. This study addresses these gaps by introducing a multi-agent framework for spatial Text-to-SQL and by developing a benchmark that emphasizes workload coverage and geometric diversity rather than scale alone.

## 3. Methodology

The methodological framework for this study is based on a multi-agent LLM system in which multiple agents cooperate to translate natural language questions into executable SQL, with particular attention to spatial reasoning. The decomposition is guided by four design principles. The first is separation by cognitive function, because spatial Text-to-SQL combines language interpretation, schema grounding, spatial reasoning, code generation, and validation. The second is grounding before generation, because later stages should operate on evidence from the schema rather than on guessed database structure. The third is progressive constraint, where each stage reduces the search space for the next one. The fourth is verification after synthesis, because a spatial query that looks plausible as text may still be invalid or semantically wrong when checked against the database engine. These principles shape both the system-level components and the multi-agent pipeline shown in Figure 2.

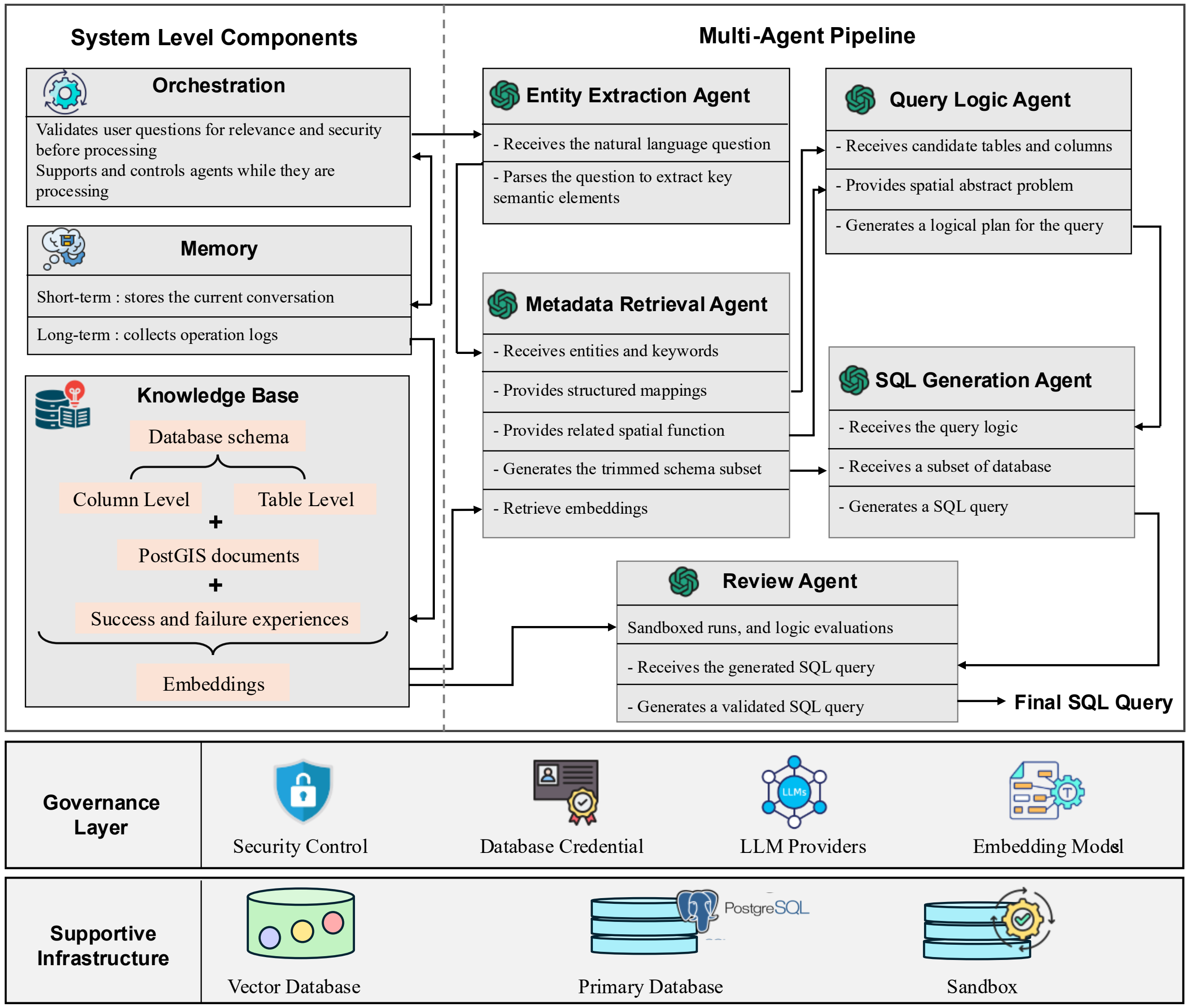

Figure 2. Architecture framework of the multi-agent spatial-to-SQL system.

### 3.1 System-Level Components

Prior to introducing the system's core concept of the multi-agent design (Section 3.2), the system-level components must be outlined, as this layer constitutes a central foundation of the multi-agent system. The system level comprises three components, including orchestration, memory component, and the knowledge base, to ensure that user queries are effectively managed, past usage are preserved, and the Text-to-SQL pipeline has access to the necessary domain knowledge.

#### 3.1.1 Orchestration

Orchestration component coordinates the overall workflow of the Text-to-SQL pipeline. In this first module, it receives a user question and determines whether the request is relevant to the primary database and if the user's intent is clear. In cooperation with the Security Control module, it also checks whether the query could result in malicious or unauthorized database access. Acting as the control unit, orchestration governs the flow of information between agents, ensuring that each one receives input in the proper format and the agent output is also correctly structured for subsequent processing. By mediating these interactions, orchestration guarantees consistency and interpretability throughout the Text-to-SQL pipeline.

### 3.1.2 Memory

Memory component is responsible for maintaining both short-term and long-term context related to queries and their results, thereby supporting coherent query generation. Short-term memory captures the state of the ongoing interaction, including the current user question and intermediate outputs from agents. This enables the system to manage multi-turn conversations where users refine, clarify, or modify their requests. After each user message, the Orchestration retrieves information from this component to interpret the user's intent, which may span multiple messages rather than a single one.

Long-term memory stores past user questions, generated queries at each step, execution results, employed functions, and user feedback. This allows the system to learn from prior tasks and interactions and improve its performance over time, enhancing both speed and efficiency by leveraging awareness of previous decisions and outcomes.

### 3.1.3 Knowledge Base

Knowledge base supports semantic reasoning by providing enriched metadata about the (spatial) database schema, which is generated and stored during system installation. Metadata generation is carried out in two phases at two levels: column level and table level. At the column level, entries capture precise counts, ranges, and structural rules (Table 1). At the table level, the Knowledge Base combines catalog checks with statistical profiling (Table 2). It identifies constraints such as primary and foreign keys, records indexed columns and row counts. When spatial or temporal fields exist, it validates geometries, computes extents, and examines time coverage.

Table 1. Column level information.

| Attribute | Description |
|---|---|
| Column Name | The identifier of the column is within its table. |
| Data Type | The PostgreSQL data type (e.g., integer, Geometry (Point,4326)) |
| Nullability | Indicates whether the column allows NULL values |
| Default Value | The default expression or constant assigned to the column, if any. |
| Foreign-Key Reference | The referenced table and column(s) that this column points to. |
| Null Count | The count of rows where this column's value is NULL. |
| Total Row Count | The total number of rows in the table when profiling this column. |
| Value-Type | A label such as "purely numeric," "purely text," "mixed type," or "numeric with string format," determined via regex and sampling. |
| Numeric Min/Max | For numeric columns, the minimum and maximum values after safe casting to numeric. |
| Unique Flag | A note indicates if every value in the column is unique. If so, the column could be used as identifier. |
| Full Unique-Values List | A complete list of all unique values when the distinct count does not exceed a configurable threshold (e.g., 1,000). |
| Sample Values | A small random selection of distinct values to illustrate typical content. |

Table 2. Table level information.

| **Attribute** | **Description** |
|---|---|
| Table Name | The identifier of the table is within the database. |
| Row Count | The total number of records on the table. |
| Column List | A comma-separated list of all column names on the table. |
| Nullable Columns | A list of column names that allow NULL values. |
| Primary Keys | Column(s) designated as the primary key constraint. |
| Foreign Keys | Mappings of each constrained column to its reference table and columns |
| Indexed Columns | Columns included in any index definitions, deduplicated. |
| Geometry Presence | A flag indicates whether the table contains a geometry column. |
| Geometry Column Details | If present, the name of the geometry column, its subtype (e.g., POINT, POLYGON), and SRID. |
| Geometry Validity | A Boolean result of ST_IsValid across all geometries in the column. |
| Spatial Extent | The bounding box of the geometry column. |
| Temporal Coverage | The earliest and latest dates/times found in any date/time-named column (e.g., "YYYY-MM-DD to YYYY-MM-DD"). |

The first phase of metadata generation is a systematic programmatic profiling of the primary database. This step extracts structural information from the database catalog including tables, columns, data types, constraints, and indexes while also computing descriptive statistics, such as value distributions, ranges, null counts, and spatial or temporal coverage. These metrics ensure agents have access to the crucial information about the primary database. The second phase leverages LLMs to translate the raw statistics into human-readable narratives that expand abbreviations, clarify semantic meaning, and contextualize values. This narrative enrichment provides interpretable, accessible summaries of schema elements for the Text-to-SQL Pipeline.

Once metadata is established, the embedding layer encodes these narratives into high-dimensional vectors using transformer-based models (Vaswani et al., 2017). This embedding process captures semantic similarities between tables and columns, enabling efficient retrieval through similarity search. When processing a user query, embeddings help identify the most relevant schema elements, which are then passed to the query generator in a concise form.

### 3.2 Multi-agent Pipeline for Text-to-SQL

The multi-agent pipeline receives a natural language query from the Orchestration component and progressively transforms it into a validated SQL statement. The workflow is not a sequence of loosely connected prompts. Instead, agents exchange typed JSON outputs that act as explicit contracts between stages. Earlier outputs are validated before downstream use, and later stages are constrained by the schema evidence and logical decisions already established. In this way, the pipeline keeps coupling and consistency across stages while still allowing each agent to specialize in a narrower task. At a high level, the Entity Extraction Agent interprets the question, the Metadata Retrieval Agent grounds it in the schema, the Query Logic Agent builds a high-level plan, the SQL Generation Agent converts that plan into executable PostGIS SQL, and the Review Agent validates and repairs the result against database behavior.

#### 3.2.1 Entity Extraction Agent

The Entity Extraction Agent is the first stage of understanding. Its role is to turn the user's natural language into a structured interpretation that later agents can trust. In spatial Text-to-SQL, this step is especially important because a place name, distance phrase, or temporal expression can play different roles in a query. The agent therefore identifies the main entities, conditions, locations, time expressions, and operation cues that describe the user's intent.

The main value of this agent is ambiguity control. It produces a structured output rather than free-form text, and it can stop early when an unresolved location, timeframe, or other blocking ambiguity would make downstream reasoning unreliable. In this sense, the Entity Extraction Agent gives the rest of the pipeline a stable semantic starting point and reduces the risk that later agents confuse user language with database structure. The agent's prompt has shown in **Appendix I**.

#### 3.2.2 Metadata Retrieval Agent

The Metadata Retrieval Agent is the bridge between human meaning and database structure. After the user intent has been decomposed, this agent asks where in the database those concepts live. This grounding step is especially important for spatial databases because geometry-bearing tables and columns are often named inconsistently, and the intended answer may require joining multiple spatial and non-spatial tables. The agent therefore maps extracted concepts to candidate tables, columns, and relevant spatial functions using schema evidence rather than guesswork.

$$cosine_similarity(q,c) = \frac{q.c}{\parallel q \parallel \parallel c \parallel} = \frac{\sum_{i=1}^{d} q_i c_i}{\sqrt{\sum_{i=1}^{d} q_i^2}\ \sqrt{\sum_{i=1}^{d} c_i^2}} \quad (1)$$

Its output is a narrowed and enriched schema view that later agents can use with higher confidence. By attaching descriptions, representative values, and geometry-related context to selected schema elements, the agent gives the planning and SQL stages a grounded data anchor. This design reduces hallucination, lowers the chance of wrong geometry-column selection, and keeps later reasoning bounded by validated schema rather than open-ended generation.

### 3.2.3 Query Logic Agent

The Query Logic Agent is the reasoning stage of the workflow. Once relevant schema elements are known, the system still needs to decide how the answer should be produced. The agent therefore transforms the grounded context into a stepwise logical plan. In spatial cases, this means making explicit decisions about operation order, join structure, filtering logic, aggregation scope, and whether a spatial function is compatible with the intended geometry types and measurement semantics.

Separating logic from syntax has two advantages. First, it reduces the risk of generating SQL too early, before the reasoning path is clear. Second, it improves transparency because the intermediate plan can be inspected if a later stage fails. The agent also trims the schema to the minimal task-specific subset, so the SQL generator works in a smaller and more reliable search space.

Box 1: Logical plan generated by the Query Logic Agent for question "Find all GHCN stations that intersect Pennsylvania."

***Tables & Columns***

**GHCN**

- geom: Role = join key. Used in spatial intersection join with states.geom.
- station_id: Role = output field. Identifier for stations found in Pennsylvania.
- name: Role = output field. Human-readable station name.
- lat: Role = unused. Not needed because spatial relation is determined by geom.
- lon: Role = unused. Same reason as lat.
- elev: Role = unused. Not required to answer which stations intersect the state.

**states**

- geom: Role = join key. Used in spatial intersection join with ghcn.geom.
- name: Role = filter criterion. Used to select the Pennsylvania polygon.

- stusps: Role = unused. Mis-coded in sample rows; not needed if filtering by name.

***Join Strategy***

- Perform a spatial INNER JOIN between ghcn and states on ST_Intersects(ghcn.geom, states.geom).
- INNER JOIN is appropriate because we only want stations that intersect the Pennsylvania geometry; stations not intersecting Pennsylvania should be excluded.

***Filter Conditions***

- states.name = 'Pennsylvania' to select the Pennsylvania state geometry.
- (If multiple Pennsylvania polygons exist, no additional filter needed; the spatial join will handle multipart or multiple rows for the same state name.)

***Output Definition***

- ghcn.station_id AS station_id: The unique station identifier to report which stations intersect Pennsylvania.
- ghcn.name AS station_name: The station's name for readability.
- Optionally, ghcn.geom AS geom: Include if geometry needs to be returned; otherwise omit for a lighter result.

***High-Level Algorithm***

- Filter states to the row(s) where name = 'Pennsylvania'.
- Spatially join ghcn to the filtered Pennsylvania geometry using ST_Intersects(ghcn.geom, states.geom).
- Select station_id and name (and geom if needed) from the matched ghcn rows.
- Return the resulting set of stations.

### 3.2.4 SQL Generation Agent

The SQL Generation Agent is the implementation stage. It takes the logical plan and the narrowed schema context and translates them into executable SQL. In this framework, SQL generation is treated as controlled translation rather than free-form writing. The agent is expected to stay within known schema boundaries, apply the chosen logical steps faithfully, and respect spatial assumptions such as CRS-aware measurements and the use of appropriate PostGIS functions.

Its output includes the SQL statement and a brief manifest of assumptions, such as literal matching choices or CRS-related interpretation. This makes the stage more transparent and makes it easier for the final reviewer to understand why the query was produced in a particular form. By locating SQL writing after interpretation, grounding, and planning, the framework reduces the chance that the model invents structure or hides reasoning mistakes inside seemingly correct syntax.

### 3.2.5 Review Agent

The Review Agent is the final quality-control stage. It treats generated SQL as something that should be tested, criticized, and revised when needed rather than trusted after one pass. This is especially important in spatial Text-to-SQL because a query may look correct as text while still failing in practice due to a wrong column, incompatible function use, or a subtle mismatch between the requested spatial meaning and the implemented SQL.

The review process is execution-aware and schema-bounded. The agent checks the query against the database environment, inspects whether the result is consistent with the user question, and attempts repair only within the grounded schema context. This stage closes the loop between language reasoning and database reality, and it turns the pipeline from a one-shot generator into a more dependable and self-correcting system. Figure 3 shows the Review Agent architecture.

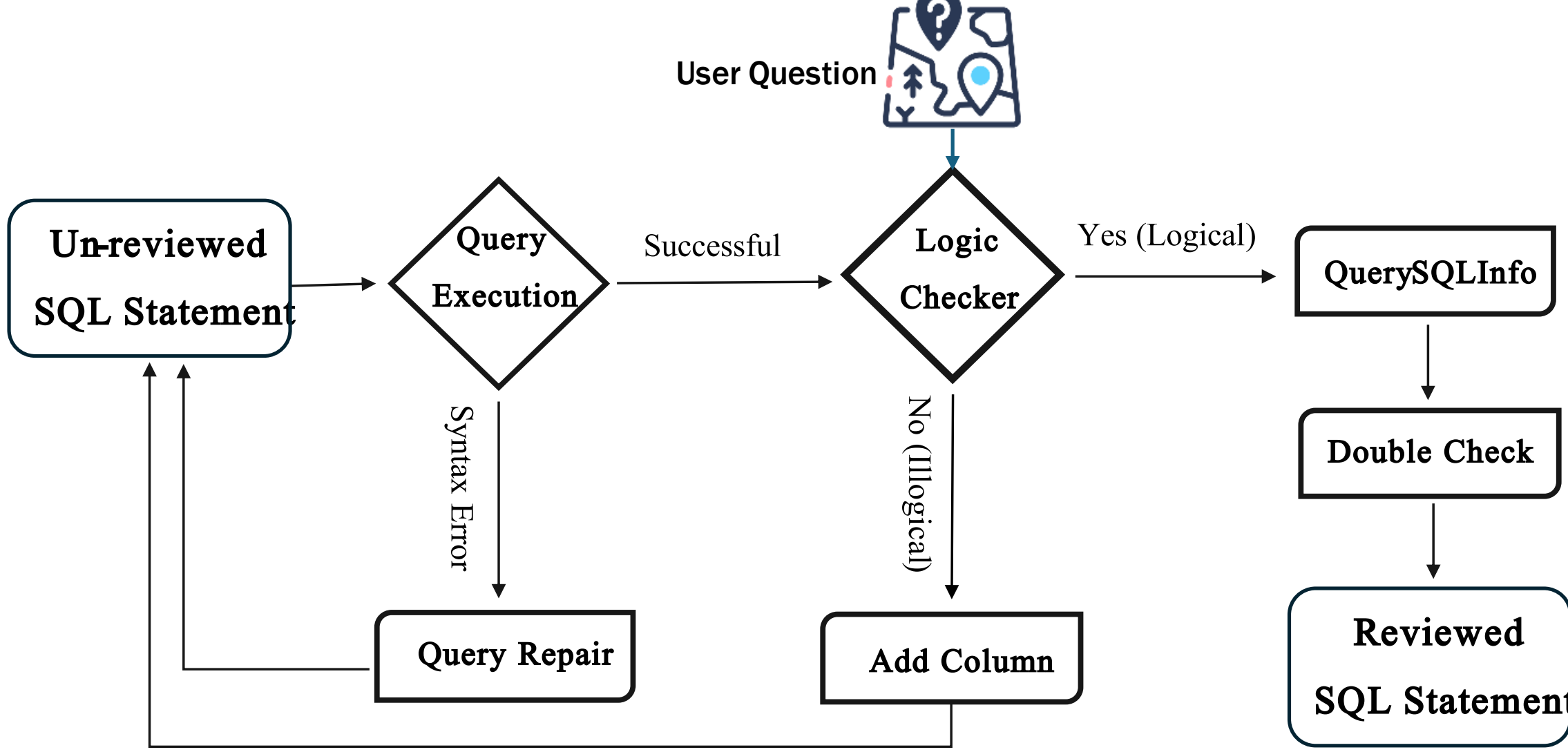

Figure 3. Workflow of the Review Agent.

The agent follows a set of PostGIS spatial query constraints, ensuring correct CRS usage, geometry or geography selection, function semantics, and spatial relationships (**Appendix II**). It enables the agent to assess accuracy, consistency, and logical integrity during spatial SQL validation.

Next, the *Review Agent* executes a sandboxed dry run using a QueryExecution tool which makes a read-only connection to the primary database with safety wrappers and appending a LIMIT 10 clause to fetch representative rows. It formats the result into a compact, human-readable table for quick inspection and hands the rows back to an LLM-assisted evaluator to confirm whether the returned sample correctly answers the user's question. If the sample indicates missing information, for example, a required column was omitted, the agent invokes its AddColumn function. This function gets the missing column(s), identifies the related column(s) in the schema description, injects the column(s) into the SELECT clause, and finally returns the revised SQL.

Beyond output inspection, the *Review Agent* has access to deeper programmatic tools. First, the QueryInfo function which parses SQL to produce a machine-readable manifest of base columns, predicates, and join operations. Second, DoubleCheck function cross-validates the SQL against natural-language schema descriptions to confirm that every referenced table and column exists, and that literal values match declared types. The DoubleCheck function also checks spatial parameters to verify coordinate reference systems, units, and appropriate spatial function usage. If any check fails or detects a semantic drift, the *Review Agent* returns a corrected SQL statement to the end-user.

Because review is separated from generation, the system gains a dedicated critic stage with a different responsibility from the creative SQL-writing stage. This separation is conceptually important. It adds humility to the workflow, acknowledges that first-pass SQL may be imperfect, and improves trust in the final answer by requiring that it survive contact with the database engine.

### 3.3 Supporting Infrastructure and Governance Layer

The supporting infrastructure is a collection of tools that enable efficient, accurate, and secure management of geospatial Text-to-SQL operations (Figure 2). It encompasses the primary PostgreSQL database for storing the (geospatial) datasets and a vector database that maintains the embeddings of database schema. A crucial component of this supporting infrastructure is the use of embeddings. Traditional database querying relies on exact keyword matching between user input and schema definitions, which often leads to mismatches when users employ alternative terminology, synonyms, or domain-specific phrases. Embeddings address this limitation by representing tables, columns, and their relationships in a high-dimensional semantic space, where similar semantically similar concepts are positioned near each other, irrespective of the exact wording used. This capability is particularly important for geospatial applications, where users may describe spatial relationships, thematic attributes, or domain-specific concepts in varied ways. For example, a user may ask for "town centers" while the schema contains "urban_centroid". Embeddings bridge this gap, enabling accurate query generation, adaptability to evolving schemas, and robust Retrieval-Augmented Generation (RAG) through relevant schema ranking. By supporting natural language interaction instead of requiring schema memorization, embeddings enhance SQL generation accuracy while reducing users' cognitive burden. The final element of the supporting infrastructure is the query execution module, which functions as a controlled sandbox environment for running automatically generated SQL statements.

The governance layer is extended to incorporate additional features, including LLM model selection, embedding model selection, security control, and database credential. It ensures that system operations remain manageable, reliable, and secure. Within this layer, configuration management centralizes the control of critical parameters such as database connections, embedding models, and LLM providers. Maintaining these configurations in a structured and consistent manner supports easier deployment, scalability, and updates, while minimizing the risk of errors from inconsistent settings. Additionally, a built-in security mechanism blocks any query that attempts to add new rows (INSERT), modify existing data (UPDATE), manage database objects (CREATE, ALTER, DROP), or delete rows (DELETE).

### 3.3.1 Model and Decoding Configuration

All agents use GPT-4o as the main generative model, while schema and metadata retrieval rely on OpenAI text-embedding-3-small for semantic search. GPT-4o was selected because the pipeline requires reliable tool use, structured outputs, multi-step reasoning, SQL generation, and SQL repair across several stages. Although smaller models can be attractive for lower-cost stages, using one main model across the pipeline improved consistency and reduced failure propagation between stages.

Decoding settings favor determinism and structured output rather than creative variation. The Entity Extraction Agent uses temperature 0.0 and a strict JSON schema to reduce parsing errors. The Metadata Retrieval Agent and Query Logic Agent rely mainly on tool-calling and schema-bounded intermediate reasoning. The SQL Generation Agent and Review Agent also use constrained output settings because their responses must be executable, precise, and easy to validate programmatically. These design choices were guided by reliability, not stylistic richness.

## 4. Evaluation Design and Benchmark Datasets

The proposed system was evaluated using spatial and non-spatial datasets, which will be discussed in detail in this section.

### 4.1 Evaluation Design

To assess the queries generated by the multi-agent system, we adopted a manual, rationale-based evaluation protocol. We did not use automatic metrics such as execution accuracy or exact string matches, as they are unsuitable for complex queries. First, many questions allow multiple correct SQL formulations, so exact match penalizes semantically correct but structurally different queries. Second, some benchmark queries (e.g., in KaggleDBQA) are debatable or incomplete. Third, LLMs often generate enhanced queries with additional columns or details for clarity, which execution accuracy would wrongly mark as errors. Since Text-to-SQL systems aim to return queries that reliably answer the user's question, we adopted a manual evaluation approach. Each query was determined as either correct (reliable and aligned with the question) or incorrect (misaligned or unreliable), and every decision/operation was accompanied by the underlying reasoning. This not only ensured that evaluation reflects semantic correctness and user intent rather than superficial similarity but also provided qualitative insight into systematic error patterns.

To evaluate the performance of the proposed system, we evaluate the queries generated before and after the involvement of *Review Agent*. Each System-Generated Query (SGQ) was compared against the Benchmark-Proposed Query (BPQ). If the SGQ matched the BPQ, it was deemed correct. Otherwise, a manual inspection was conducted to determine why the SGQ differed. If SGQ execution produced expected results despite not structurally matching the BPQ, it was still considered correct. In cases where the BPQ itself was incorrect, the SGQ was accepted as correct if it was validated by us.

To investigate the system's performance after involvement of the *Review Agent*, the queries generated were evaluated using the same criteria as for the SGQ. First, each reviewer-generated query was compared with the unreviewed system-generated query. If the two matched, one of them was executed and its result was examined against the input question. If the result was correct, the queries were deemed correct. Conversely, if the test failed and returned unexpected output, the reviewed query was marked as incorrect. However, if the reviewed and unreviewed queries differed, a human annotator examined the differences. The annotator investigated whether the reviewed query accurately captured the intent of the original input question. Based on this judgment, the query was labeled either correct or incorrect. These evaluation procedures were applied to both spatial and non-spatial queries to determine the extent to which the *Review Agent* enhanced the system's performance.

### 4.2 Benchmarking Datasets

#### 4.2.1 Spatial Query Benchmark

We developed as a benchmark (SpatialQueryQA) with three levels of complexity including basic, intermediate, and advanced. It incorporates all types of geographical features (point, polyline, and polygon) and consists of 9 benchmark tables, each containing a geometry column. The data are derived from OpenStreetMap, the National Centers for Environmental Information database, and the United States Census Bureau (Table 3). The geometry columns of the dataset cover diverse spatial extents.

Table 3. Geospatial data sources for the SpatialQueryQA benchmark dataset.

| **Source** | **Table** | **Extent** | **Geometry type** | **Column Counts** | **Record Counts** |
|---|---|---|---|---|---|
| Open Street Map | POI | Pennsylvania, U.S. | Point | 6 | 61,665 |
| Open Street Map | Roads | Pennsylvania, U.S. | Polyline | 12 | 1,653,169 |
| National Centers for Environmental Information | Global historical climatology network | Worldwide | Point | 17 | 1,701 |
| The U.S. Census Bureau | Census block groups | U.S. | Polygon | 5 | 242,748 |
| The U.S. Census Bureau | Census tracts | U.S. | Polygon | 3 | 85,503 |
| The U.S. Census Bureau | Counties | U.S. | Polygon | 5 | 3,235 |
| The U.S. Census Bureau | States | U.S. | Polygon | 17 | 56 |
| Natural Earth Data | Protected areas | Worldwide | Polygon | 10 | 61 |
| Natural Earth Data | Time zones | Worldwide | Polygon | 17 | 120 |

*EPSG:4326 is the CRS.*

In the basic-level tasks, the dataset includes operations such as selection, filtering, area calculation, distance calculation, geometry retrieval, and attribute retrieval. Most of these are one-step operations and the system is expected to generate a single step SQL statement. At the second level of complexity, the dataset focuses on more advanced tasks, including spatial joins and topological relationships (e.g., within, intersect, overlap), as well as spatial proximity and attribute retrieval queries that require more than one step. In advanced level, the benchmark primarily focuses on aggregation and quantitative analysis (e.g., counts, averages, maxima/minima) combined with spatial operations such as containment, distance, and intersection. For this level, the system is required to generate a multiple step SQL statement to accurately retrieve data from the corresponding table (s). Each difficulty level in the benchmark includes 30 queries covering a wide range of geographic operations.

SpatialQueryQA was designed as a coverage-oriented benchmark rather than a large benchmark. When characterized by workload, the 90 questions cover topological relationships and spatial joins (31), k-nearest neighbor or distance-ranking tasks (20), attribute-based filtering on spatial tables (18), spatial measurement tasks (18), geometry attribute extraction (6), geometry processing or transformation tasks (6), and proximity or radius-search tasks (4). The total exceeds 90 because some questions involve more than one workload. The most frequent PostGIS predicates, functions, and operators include ST_Intersects, ST_Within, ST_DWithin, ST_Area, ST_Length, ST_Perimeter, ST_Centroid, ST_Buffer, ST_Union, ST_X, ST_Y, ST_GeometryType, ST_AsText, and the kNN operator <->. This characterization aligns the benchmark with practical workload families commonly discussed in spatial natural-language interfaces, including range, nearest-neighbor, spatial join, and aggregation queries (Wang et al., 2018).

### 4.2.2 KaggleDBQA

The KaggleDBQA is used as the non-spatial query test, which is a cross-domain Text-to-SQL benchmark created to evaluate semantic parsing in realistic settings (Lee et al., 2021). The collection comprises eight databases and 272 test instances that reflect substantial schema complexity and real-world heterogeneity (Table 4).

Table 4. KaggleDBQA information.

| Database name | Table Name | Column Counts | Record Counts |
|---|---|---|---|
| WorldSoccerDataBase | betfront | 11 | 27,853 |
| | football_data | 26 | 179,571 |
| Pesticide | resultsdata15 | 16 | 2,333,911 |
| | sampledata15 | 18 | 10,187 |
| USWildFires | fires | 19 | 1,880,465 |
| GeoNuclearData | nuclear_power_plants | 14 | 788 |
| WhatCDHipHop | torrents | 7 | 75,719 |
| | tags | 3 | 161,283 |
| TheHistoryofBaseball | hall_of_fame | 9 | 4,120 |
| | player_award | 6 | 6,078 |
| | player_award_vote | 7 | 6,795 |
| | salary | 5 | 25,575 |
| | player | 17 | 18,846 |
| StudentMathScore | finrev_fed_17 | 8 | 14,306 |
| | ndecoreexcel_math_grade8 | 4 | 53 |
| | finrev_fed_key_17 | 3 | 51 |
| GreaterManchesterCrime | greatermanchestercrime | 6 | 5,000 |

## 5. Experiments and Results

This section evaluates the system's performance in generating SQL statements. As outlined in the methodology, the Orchestration component acts as the main interface with users, engaging in multi-turn dialogues to clarify ambiguous inputs. In real-world use, such interactions are crucial before activating the Text-to-SQL pipeline. We first illustrate this capability through an example. Next, we evaluate the Text-to-SQL pipeline using the two selected benchmarks. Because the Review Agent has the greatest influence on the final output, we report the percentage of correct SQL statements both before (unreviewed) and after (reviewed) its involvement.

### 5.1 Multi-turn Conversations

The Orchestration does not invoke the Text-to-SQL pipeline until it has a clear understanding of the user's intent. The Memory component supplies the Orchestration with previous interactions between the user and the system, enabling it to interpret the user's intent within the broader context of an ongoing conversation rather than a single message. The Orchestration asks clarifying questions, if needed, to clearly understand the user's intent. Box 2 illustrates how the system identifies user intent in a multi-turn conversation.

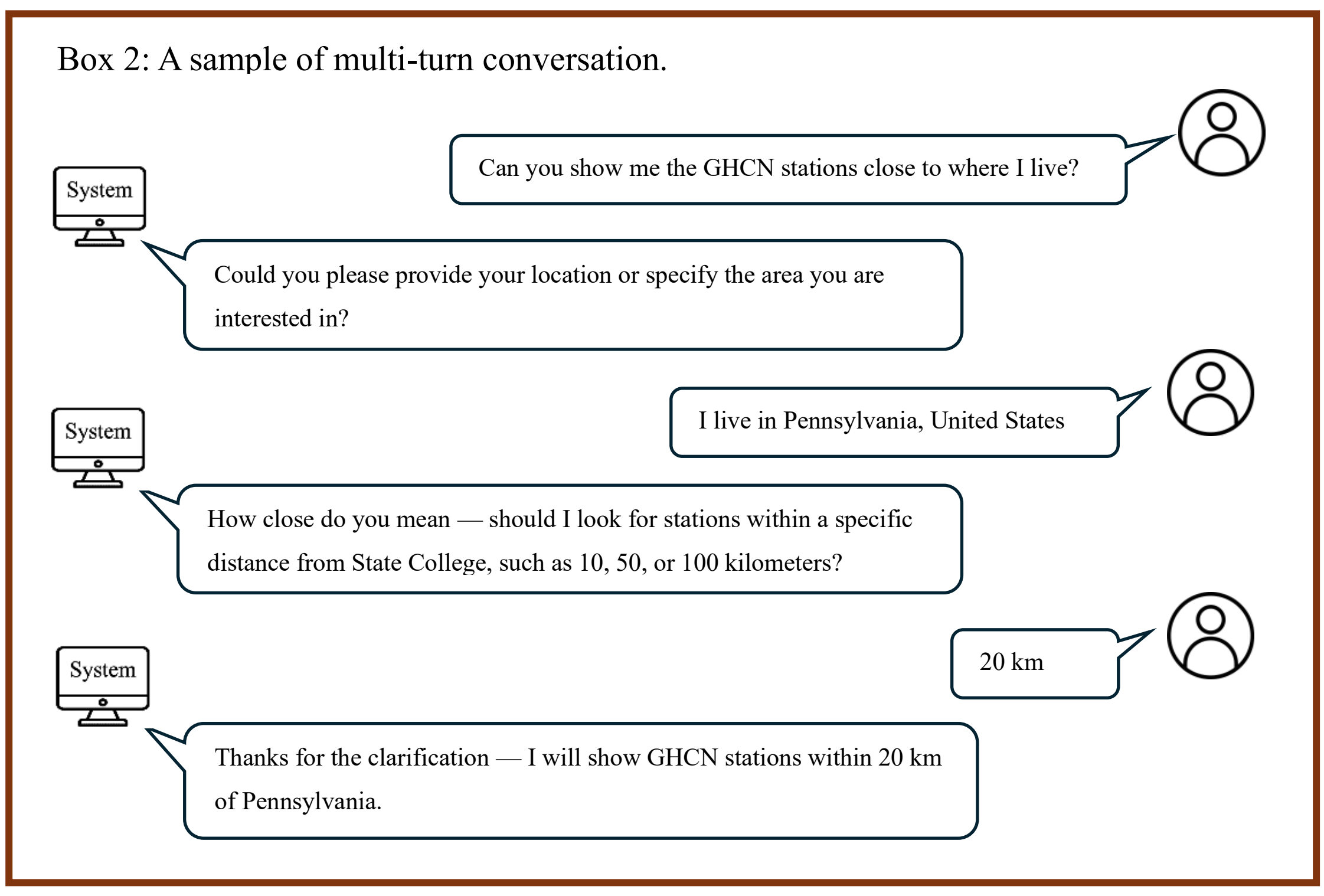

Box 2: A sample of multi-turn conversation.

This example illustrates a dynamic clarification strategy rather than a full trip-planning style conversational search. The Orchestration and Memory components work together to identify missing constraints, ask only when a blocking ambiguity is present, and delay the Text-to-SQL pipeline until the user intent is sufficiently clear. Although Box 2 mainly shows ambiguity removal, the same memory mechanism can accumulate constraints across turns and reformulate the final SQL from the combined conversation history. This interaction style is different from incremental trip-planning systems that execute a search after each user action, but it still supports multi-turn reasoning through conversational state tracking and query reformulation.

### 5.2 Non-spatial SQL Evaluation

For non-spatial queries, the system was evaluated across all eight databases in the KaggleDBQA benchmark. These databases cover a diverse range of tasks, including ranking, counting, filtering, max/min, descriptive statistics, categorical queries, and temporal analysis. All queries in the benchmark were executed by the system and subsequently evaluated manually to determine how many of the generated queries were correct and how many were incorrect. The detailed evaluation steps are provided in Section 4.1. For example, two cases are demonstrated from 'WorldSoccerDataBase' and 'USWildFires' databases (Boxes 3 and 4). Although structurally different from the benchmark queries, the system's queries produced the same results.

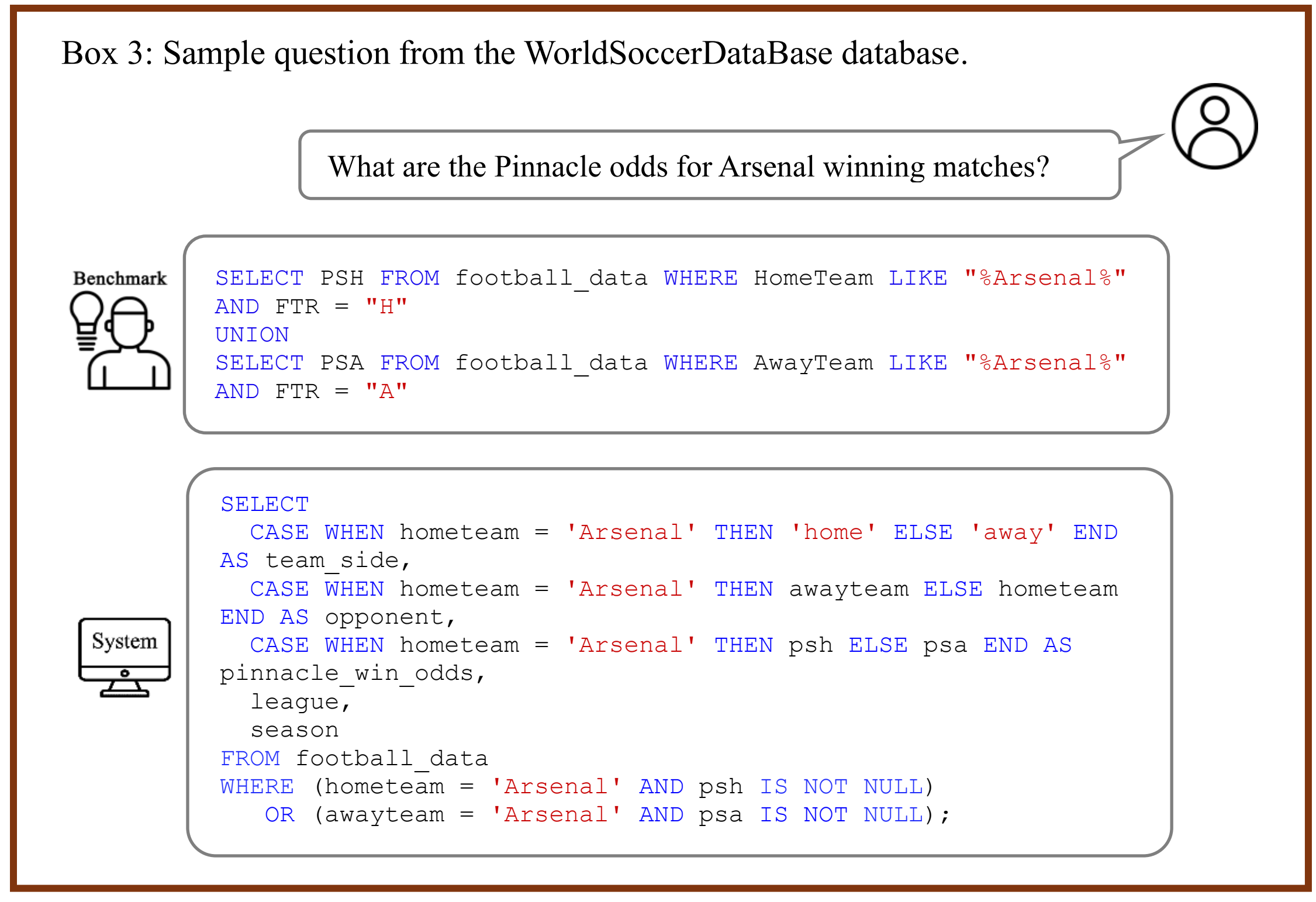

Box 3: Sample question from the WorldSoccerDataBase database.

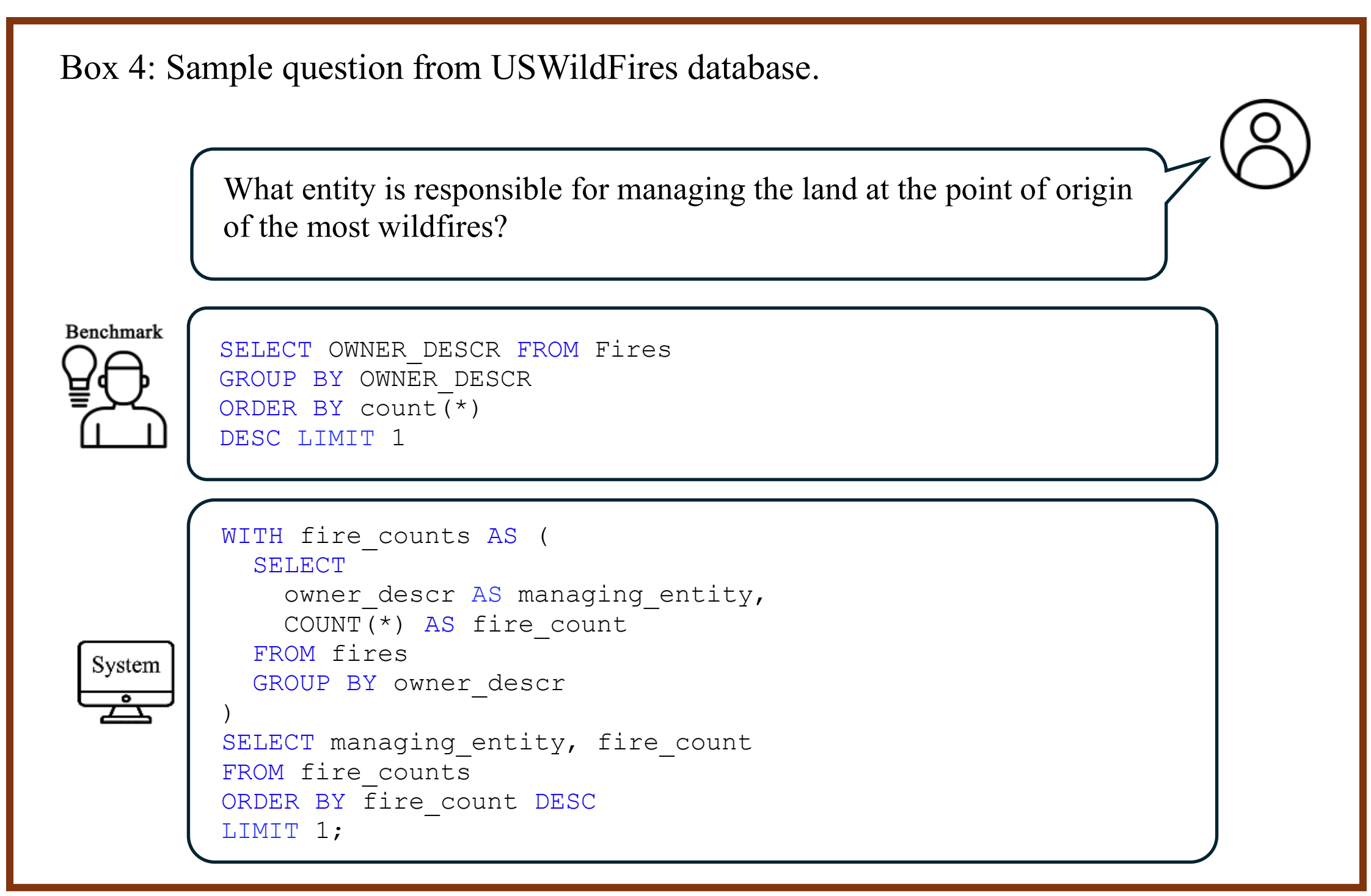

Table 5 presents results for eight non-spatial databases. The system achieved 81.2% accuracy (221/272), with performance ranging from 64.2% on *StudentMathScore* to 90.6% on *GeoNuclearData*. Detailed results and analyses for all 272 queries are in **Appendix IV**. For the same dataset, Yu et al. reported an accuracy of 56.2% for generated queries; however, it should be noted that the study has employed an execution match strategy to evaluate that is different from the employed evaluation method of this study (C. Yu et al., 2025).

Table 5. Evaluation result on non-spatial queries.

| Database | Questions Count | Unreviewed Correct Count | Reviewed Correct Count | Unreviewed Accuracy | Reviewed Accuracy |
|---|---|---|---|---|---|
| WorldSoccerDataBase | 18 | 14 | 16 | 77.8% | 88.9% |
| Pesticide | 50 | 35 | 40 | 70.0% | 80.0% |
| USWildFires | 37 | 30 | 33 | 81.0% | 89.2% |
| GeoNuclearData | 32 | 22 | 29 | 68.7% | 90.6% |
| WhatCDHipHop | 41 | 32 | 36 | 78.0% | 87.8% |
| TheHistoryofBaseball | 39 | 23 | 28 | 59.0% | 71.8% |
| StudentMathScore | 28 | 13 | 18 | 46.4% | 64.2% |
| GreaterManchesterCrime | 27 | 18 | 21 | 66.6% | 77.7% |
| **Overall** | **272** | **187** | **221** | **68.7%** | **81.2%** |

The *Review Agent* contributed a consistent accuracy gain across all datasets, improving overall performance by 12.5 percentage points (from 68.7% to 81.2%) by raising the number of correctly generated queries from 187 to 221 out of 272. Box 5 provides an example illustrating how the initial system-generated query was incorrect but was corrected by the *Review Agent*. Before the *Review Agent's* involvement, the system produced an incorrect query that returned only the numeric maximum score and limited results to 2017, which wasn't requested. After validation, the *Review Agent* correctly computed the average math score per state and identified the state with the highest average.

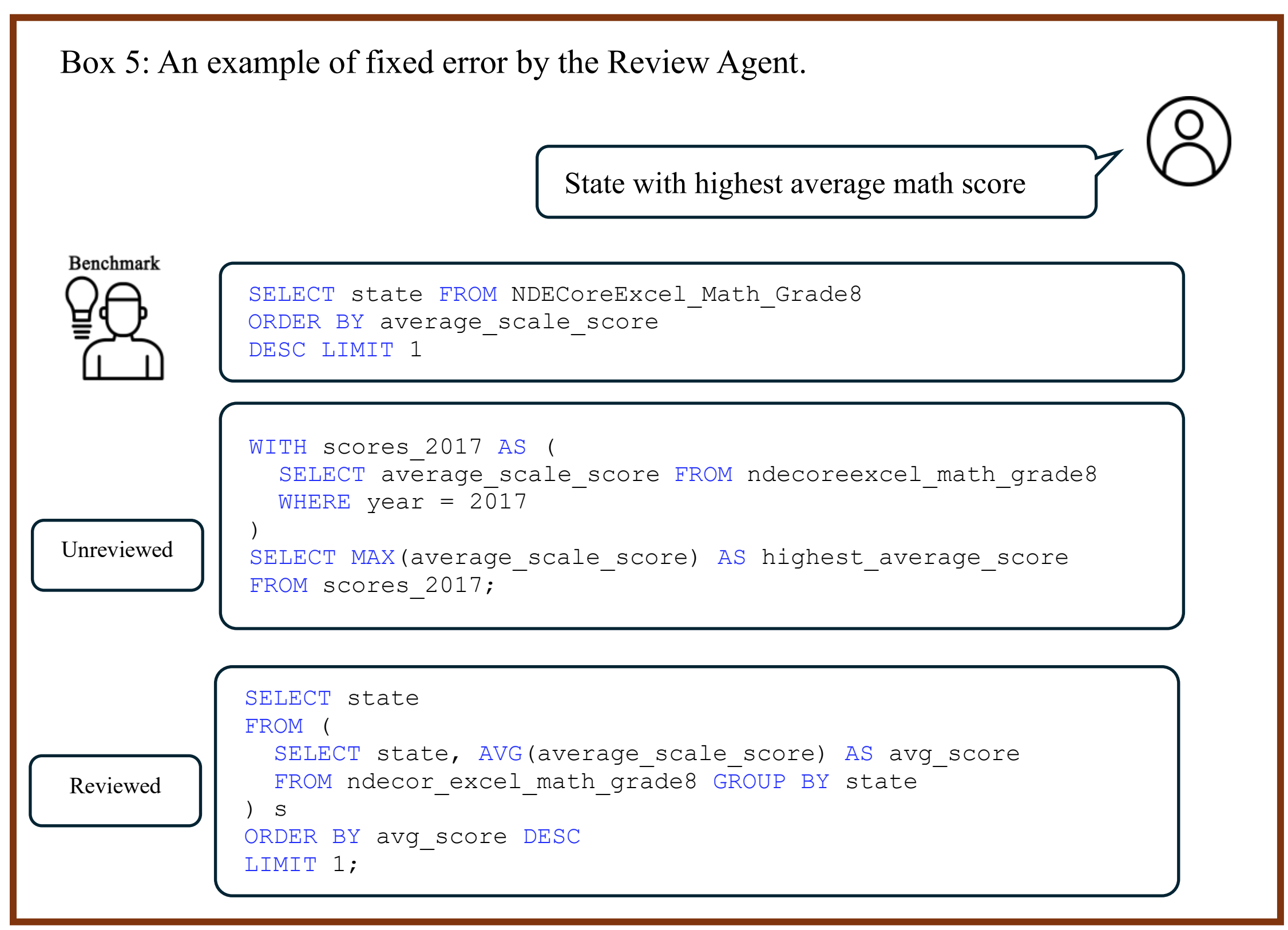

Box 5: An example of fixed error by the Review Agent.

### 5.3 Spatial SQL Evaluation

To complement the multi-agent evaluation, we also conducted a focused comparison with a single-agent baseline on 15 representative SpatialQueryQA questions, including 5 basic, 5 intermediate, and 5 advanced cases. The comparison shows a clear pattern across difficulty levels. At the basic level, the single-agent baseline returned expected results in 4 out of 5 cases. At the intermediate level, it returned expected results in only 2 out of 5 cases. At the advanced level, all 5 cases produced unexpected results. These findings reinforce the main result of this study: a single-agent approach can be adequate for simple spatial Text-to-SQL tasks, but it is not sufficient for complex queries that require tighter schema grounding, multi-step planning, correct aggregation logic, and post-generation validation. The full case-by-case analysis, including generated single-agent SQL, failure explanations, token usage, and latency, is provided in **Appendix V**.

For SpatialQueryQA, the multi-agent system achieved an overall accuracy of 87.7% (79 out of 90 queries), with accuracy of 93.3% for basic level, 90.0% for intermediate level, and 80.0% for advanced level (Table 6). The *Review Agent* again consistently improved its performance across all levels, increasing overall accuracy by 11.0 percentage points (from 76.7% to 87.7%). The largest improvement was observed in the advanced category (+13.3%). These results indicate that the reviewer plays a particularly valuable role in refining query accuracy for more complex spatial reasoning tasks. Detailed results for all 90 benchmark spatial queries are provided in **Appendix III**.

We also measured the operational overhead of the pipeline across the three difficulty levels. On average, the framework used about 11k tokens and 12 seconds per query for the basic level, about 16k tokens and 21 seconds for the intermediate level, and about 20k tokens and 27 seconds for the advanced level. The increase is expected because harder questions usually trigger more schema retrieval, more reasoning steps in the Query Logic Agent, and a higher chance that the Review Agent performs additional validation or repair. These results make the accuracy-cost trade-off more explicit. The multi-agent design adds inference overhead, but the growth is closely related to task complexity and remained within an interactive range for the evaluated benchmark.

Table 6. Evaluation results on spatial queries for different complexity levels.

| Difficulty Level | Questions Count | Unreviewed Correct Count | Reviewed Correct Count | Unreviewed Accuracy | Reviewed Accuracy |
|---|---|---|---|---|---|
| Basic | 30 | 25 | 28 | 83.3% | 93.3% |
| Intermediate | 30 | 24 | 27 | 80.0% | 90.0% |
| Advanced | 30 | 20 | 24 | 66.7% | 80.0% |
| **Overall** | **90** | **69** | **79** | **76.7%** | **87.7%** |

### 5.3.1 Basic Level Cases

As shown in Table 6, the system performs well at basic level, both with and without the *Review Agent*. Queries at this level were designed to require only one or two straightforward operations. For example, when asked to identify the coordinates of a weather station with a specific ID, the system first located the corresponding table (ghcn), filtered the rows to match the given station ID, and then returned the longitude and latitude of that station. Box 6 presents a case showing the generated SQL before and after the Review Agent. Both queries are correct and yield the same result despite structural differences.

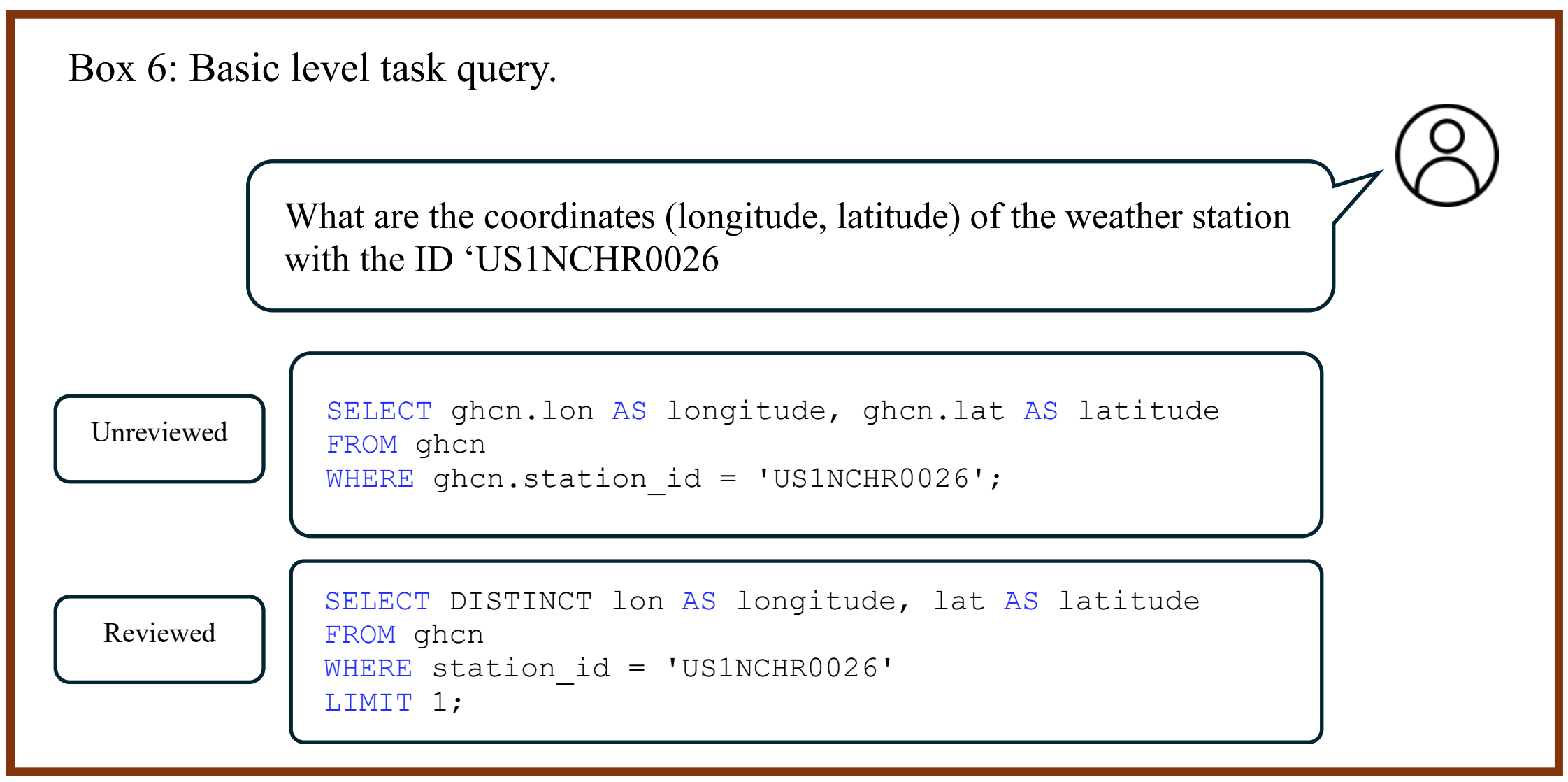

Box 6: Basic level task query.

Box 7 illustrates a case where the unreviewed SQL statement was incorrect, but the *Review Agent* successfully corrected the error. In this case, the system was tasked with calculating the area of the protected region 'Everglades' in square kilometers. The unreviewed query led to error because it filtered strictly on *unit_name = 'Everglades'*, which was likely too restrictive, and it did not apply aggregation, potentially returning multiple rows instead of a single total area. In contrast, the reviewed query produced the correct result by summing all matching geometries, computing the geodesic area through the geography type, and applying a case-insensitive partial match to more realistic names such as 'Everglades National Park'.

As illustrated, the two SQL statements differ between these stages. The *Review Agent* functions as a self-verifying component: it approves a query if it is correct, and if the SQL statement appears incorrect, it redirects the system to revise and regenerate the query. At the basic level, errors decreased from 5 to 2 after *Review Agent's* involvement.

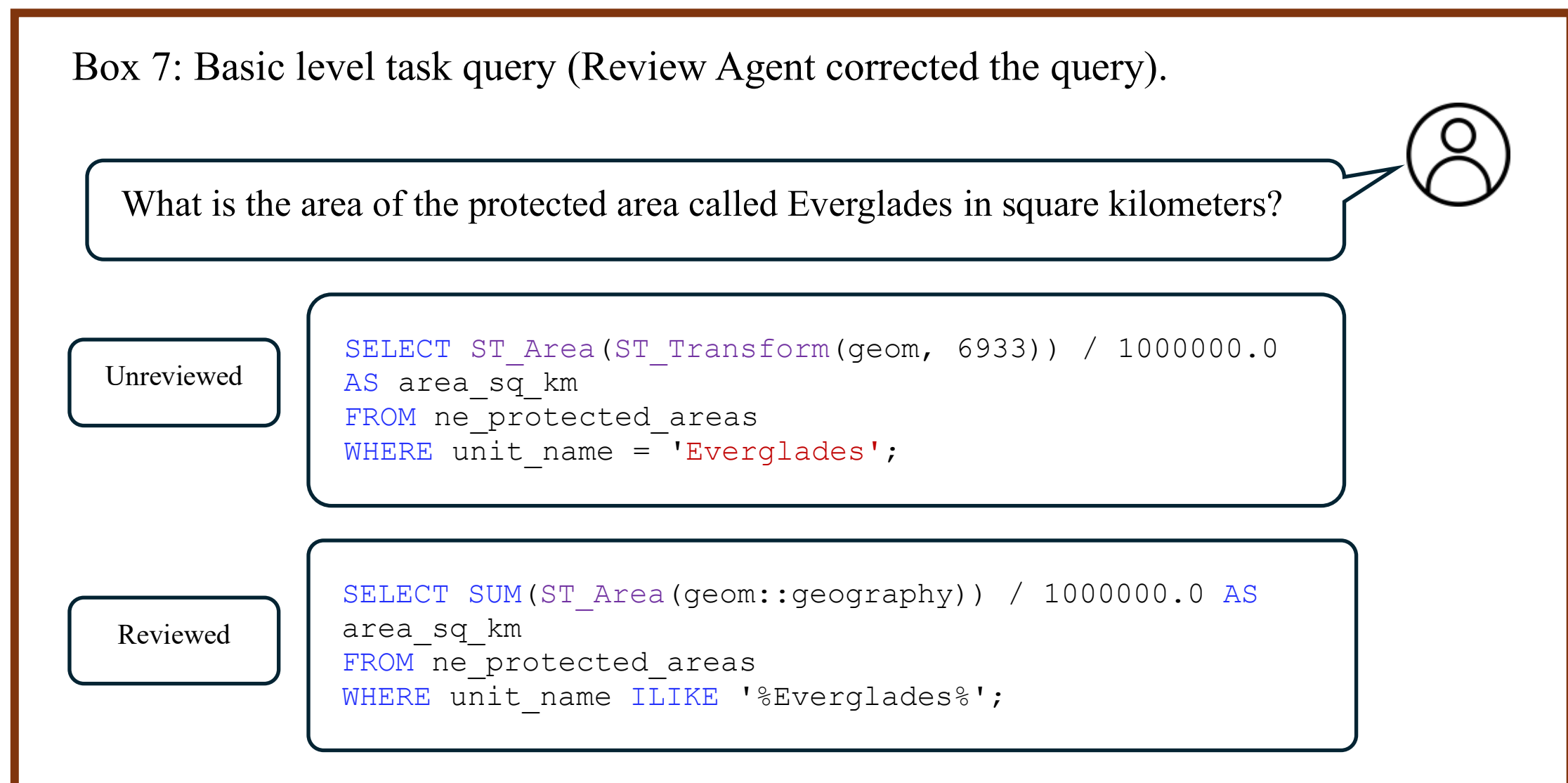

Box 7: Basic level task query (Review Agent corrected the query).

### 5.3.2 Intermediate Level Cases

The intermediate cases were designed with a higher level of complexity than the basic ones. At this level, queries typically require at least two or more steps. We illustrate this with two examples: one that resulted in a correct query and another that produced an error. In the successful case, the system was asked to identify all census tracts intersecting with the county of "Conecuh" and it accurately generated the corresponding SQL statement (Box 8). Both the unreviewed and reviewed SQL statements are identical and correct.

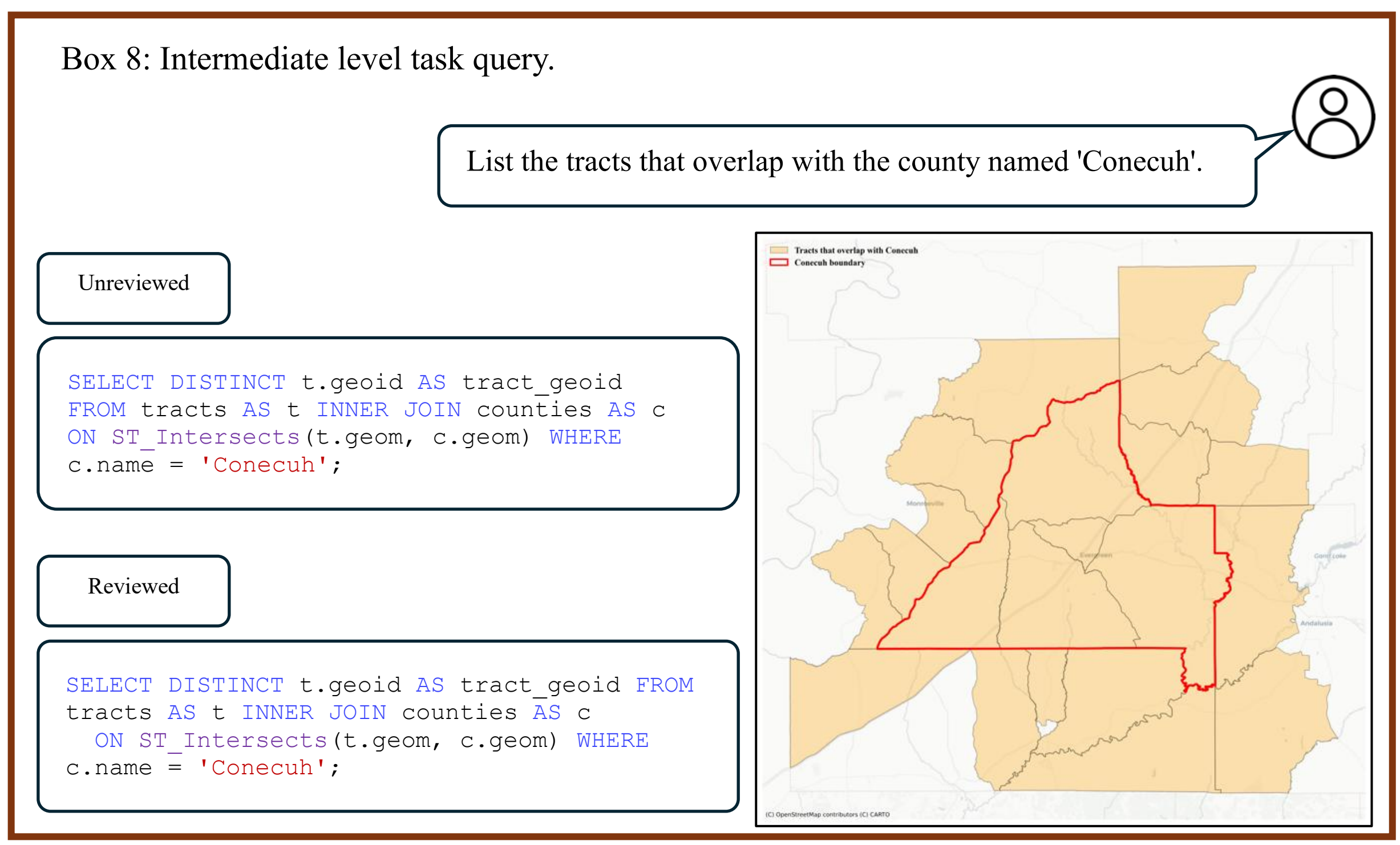

The second case illustrates a scenario where the system failed to generate the correct SQL statement. The query produced by the system in response to the question was incorrect because it calculated the distance using geometries in EPSG:3857. This projection introduces planar distortions, which are particularly problematic near the dateline where the '+14' time zone is located. As a result, some stations could be misclassified in relation to the 10-kilometer threshold. More details about Level 2 cases can be found in **Appendix III** (Level 2). At the intermediate level, errors dropped from 6 to 3 after *Review Agent* Involvement.

### 5.3.3 Advanced Level Cases

The advanced-level queries predominantly fall within the range of 3 to 5 steps, with several requiring more than 5 steps. The system must perform multi-step reasoning that extends well beyond a straightforward reading of the question. For instance, the system was tasked to examine counties within each state, measure the length of their boundaries in kilometers, and then compute the average of those perimeter values. Since both queries were identical, only one SQL statement is shown in Box 9 with its output on the right.

Box 9: Advanced level task query.

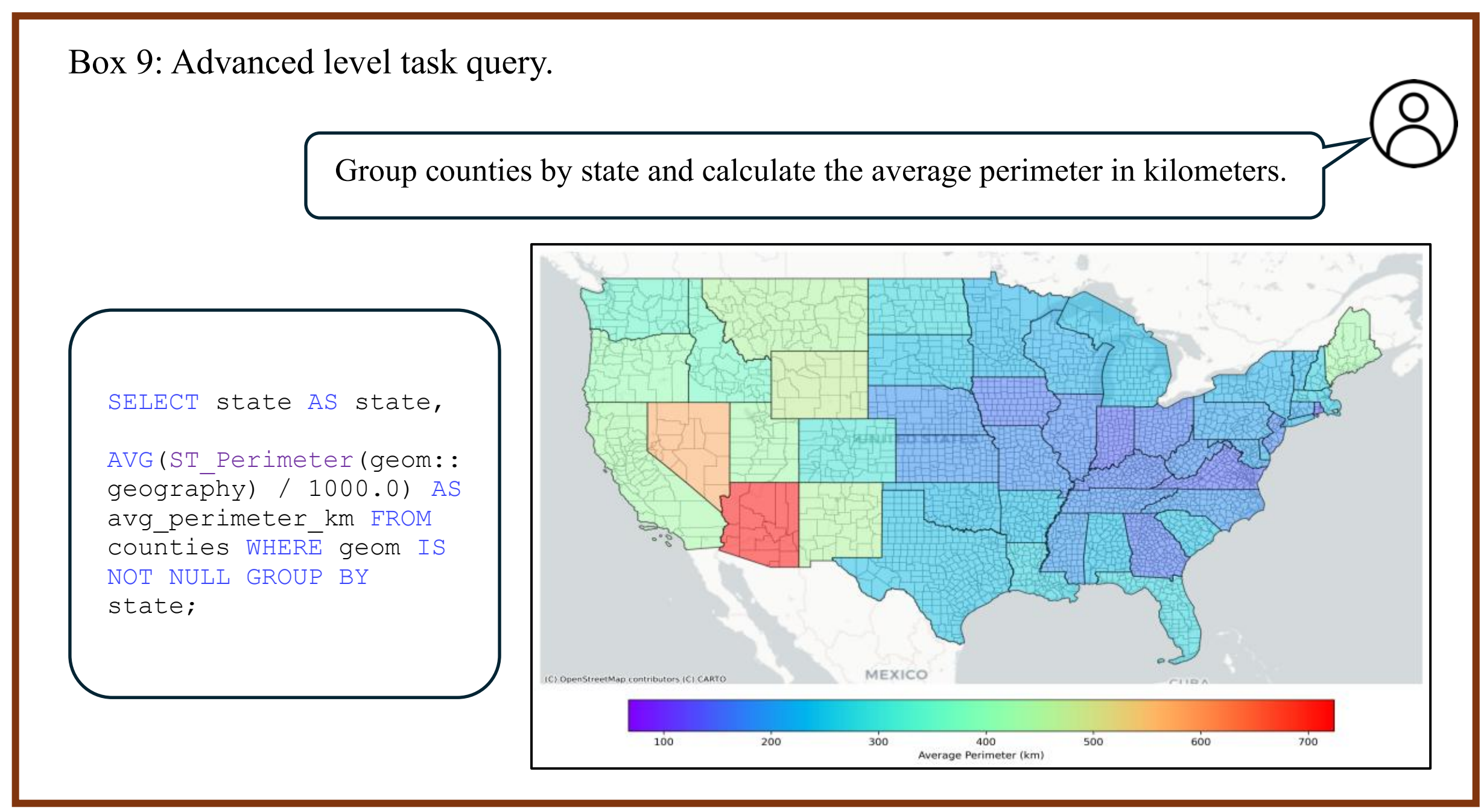

The second case demonstrates a scenario where the system-generated query was incorrect prior to the *Review Agent's* involvement but was corrected during review process. The task was to identify the WGS 84 for the time zone where "New Zealand" appears in the place column. In the unreviewed version, the query used an incorrect table name and applied a filter on the time zone column instead of the place column specified in the question (Box 10). The *Review Agent* correctly used the *place* column instead of *time zone* (Box 10), with the output shown on the right.

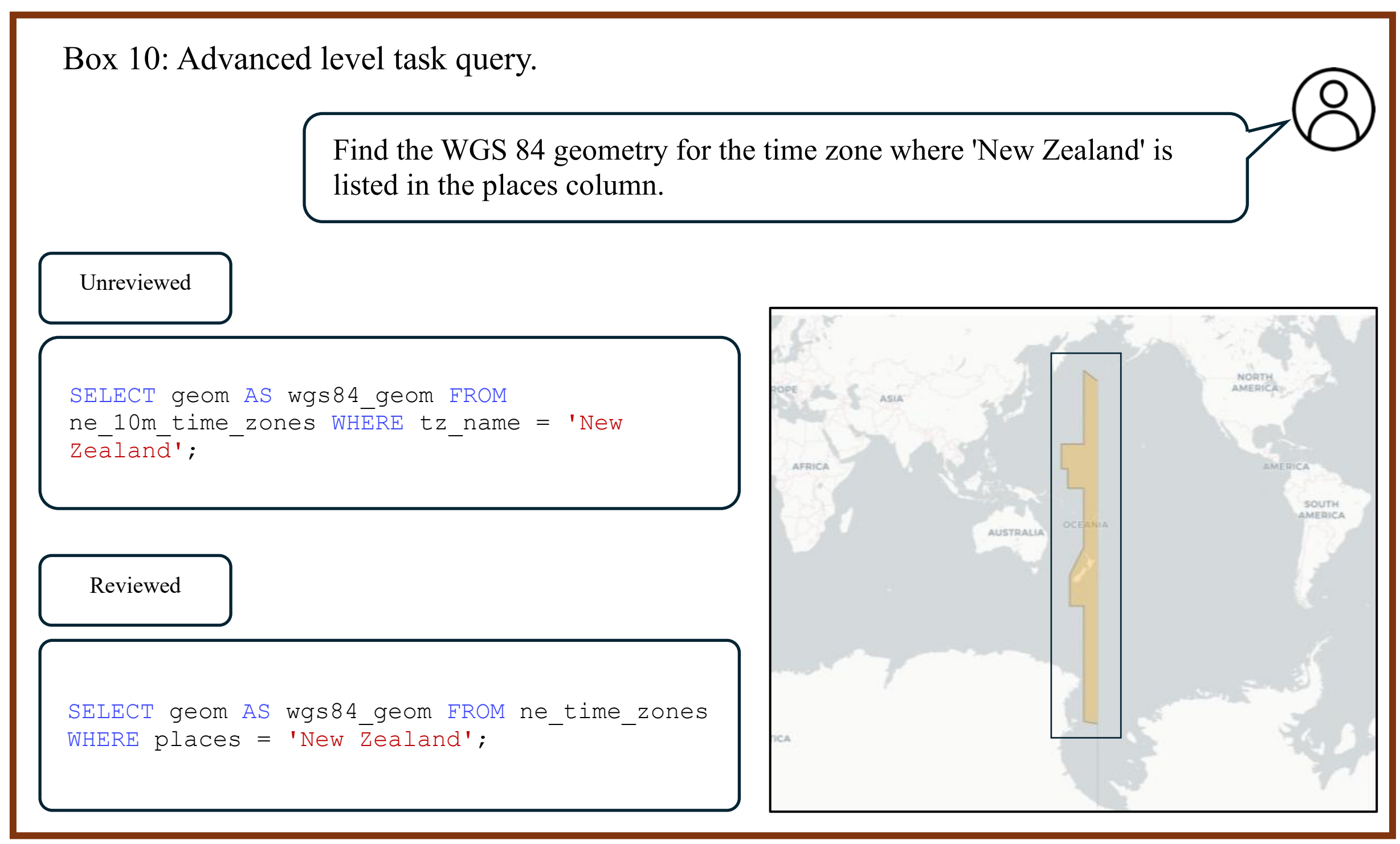

Box 10: Advanced level task query.

For the third case, the system was unable to generate a correct SQL statement, even with the *Review Agent*. In this task, the system was asked to calculate the area of all block groups in square meters that intersect with multiple census tracts. Although this task appears straightforward, it depends on a precise order of operations: first identify block groups that intersect more than one tract, then calculate their areas in square meters. It involves four steps (1) spatial join, (2) count overlaps, (3) filter to keep block groups with more than one intersecting tract, and (4) area calculation. The system-generated query performed the first three steps correctly but returned per-block-group areas instead of the single total area requested; it omitted the final aggregation (e.g., SUM). The system-generated SQL and the proposed correction are provided in **Appendix III** (Level 3). Overall, errors dropped from 10 to 6 after the *Review Agent's* involvement.

### 5.4 Self-improving

Self-improving (or self-growing) is a critical capability of autonomous systems (Li & Ning, 2023b), therefore, the multi-agent system implemented this function by recording prior interactions, generated outputs, along with user feedback to enable continual learning. While it does not explicitly analyze patterns in past errors or successes in the current implementation, it references similar previous questions to avoid repeating mistakes and to align with outcomes that were previously satisfactory.

Box 11 illustrates how the system improves SQL generation by learning from previous errors. When asked to find POIs within 5 km of each county centroid in Pennsylvania, the first query misused ST_DWithin by using a Boolean condition (< 5000) instead of specifying the distance argument, causing an error. This mistake was stored in memory, and in the next attempt, the system correctly used 5000 as the third argument to ST_DWithin.

Box 11: Find all POIs within 5 km of each county centroid in Pennsylvania..

**First run**

```sql
SELECT
    c.name AS county_name,
    p.name AS poi_name,
    p.fclass AS poi_type,
ST_Distance(ST_Centroid(c.geom)::geography,
p.geom::geography) / 1000 AS distance_km
FROM
    counties AS c
JOIN
    poi AS p
ON
    ST_DWithin(
        ST_Centroid(c.geom)::geography,
        p.geom::geography
    ) < 5000
WHERE
    c.state = '42'
    AND c.geom IS NOT NULL
    AND p.geom IS NOT NULL
ORDER BY
    c.name, distance_km;
```

**Second run**

```sql
SELECT
    c.name AS county_name,
    p.name AS poi_name,
    p.fclass AS poi_type,
    ST_Distance(ST_Centroid(c.geom)::geography,
p.geom::geography) / 1000 AS distance_km
FROM
    counties AS c
JOIN
    poi AS p
ON
    ST_DWithin(
        ST_Centroid(c.geom)::geography,
        p.geom::geography,
        5000)

WHERE
    c.state = '42'
    AND c.geom IS NOT NULL
    AND p.geom IS NOT NULL
ORDER BY
    c.name, distance_km;
```

## 6. Discussion and Lessons Learned

This study represents a significant step toward the realization of autonomous GIS (Li et al., 2023), concretely implementing several of its core goals through a multi-agent, AI-powered framework. Our system embodies the "self-generating" and "self-executing" principles by autonomously producing and running SQL queries from natural language. The integration of the *Review Agent* demonstrates the "self-verifying" goal, a key capability for building trustworthy autonomous systems. In addition, the system implements the "self-improving" principle through its Memory component, which retains both short-term and long-term records of previous interactions. By referencing these memories, the system continuously improves over interactions, avoiding repeated errors and aligning outputs with previously satisfactory results. The demonstrated performance, where the system not only matches but, in some cases, surpasses benchmark-proposed queries, shows the potential capabilities of AI to act as the core of an "artificial geospatial analyst". By successfully decomposing complex spatial questions into logical plans and executable code, while learning from past experiences, the multi-agent system provides a valuable reference for automating geospatial data retrieval and analysis, thereby lowering the technical barrier and making spatial databases accessible to a broader audience.

At the same time, our findings do not suggest that a multi-agent framework is always required. Simpler questions that mainly involve direct attribute filtering or straightforward spatial predicates may be handled adequately by a strong single-agent system, often with lower latency and lower token cost. During development, our pilot single-agent tests were usually faster, but they were less reliable on cases that combined schema ambiguity, implicit spatial operations, CRS-sensitive measurement, multi-step aggregation, and post-generation repair. We therefore frame the proposed architecture as a robustness-oriented design for compound spatial cases, not as a universal replacement for simpler approaches. A more systematic ablation against single-agent, chain-of-thought, and in-context-learning variants remains an important direction for future work. A focused comparative summary with a single-agent baseline is provided in **Appendix V**, and it shows that performance declines sharply as the spatial reasoning and SQL composition become more complex.

Despite its promising results, our evaluation reveals several key limitations that highlight the challenges on the path to full autonomy. The primary issue lies in geometric reasoning. The system occasionally fails to use correct geodesic distance calculations, introducing errors by measuring in planar projections (e.g., EPSG:3857) instead of geographic coordinates. Similarly, it can misinterpret geometric operations, such as using ST_Boundary when the full polygon geometry was intended. At advanced levels of complexity, the system struggles with precise aggregation semantics, sometimes returning per-feature results instead of a total sum.

These missteps highlight the challenge of encoding the vast and often implicit knowledge of geographic data models and domain expertise into an AI system. The discrepancy between our system's outputs and some benchmark-proposed queries also points to a broader issue: the quality and consistency of existing benchmarks themselves, which can inherit errors or suboptimal practices from their human creators.

These limitations provide a clear agenda for future research to advance the capabilities of autonomous spatial Text-to-SQL systems. First, the development of dedicated spatial reasoning modules is crucial. These modules would enforce correct spatial measurements (geodesic vs. planar), validate geometry types, and ensure appropriate use of spatial functions, directly addressing the most common spatial errors. Second, to handle ambiguity, future systems should incorporate interactive and dynamic prompting strategies. When user intent is unclear such as "whether to return boundaries or full polygons" the system should proactively ask the user for clarification, creating a collaborative human-AI problem-solving loop. Of course, the clarification questions should be generated not only before the beginning of the procedure but also in each step of the process. Third, robustness can be enhanced by building a library of dataset-specific cleaning rules and conventions. This would involve automated procedures for trimming and casting textual numeric, normalizing missing-value representations, and understanding common schema naming patterns, thereby reducing errors arising from data heterogeneity. Finally, although we have proposed a spatial query QA benchmark in this study, our findings call for a community-wide effort to develop diverse benchmarks and improve the design of available benchmarks. Future benchmarks should be rigorously validated to ensure that proposed queries reflect best practices for accuracy, robustness, and reproducibility. By addressing these frontiers, we can further close the gap between intuitive natural language interaction and the powerful data retrieval and analysis enabled by spatial SQL, accelerating progress towards Autonomous GIS (Li and Ning et al., 2025).

## 7. Conclusion

We designed, implemented, and evaluated a novel multi-agent framework to address the complex challenge of translating natural language questions into accurate spatial SQL queries. By moving beyond single-agent prompt engineering, our framework leverages a collaborative ecosystem of specialized agents that each agent is responsible for distinct tasks, from entity extraction and semantic schema retrieval to logical planning and code generation, to make geospatial databases accessible for non-experts. The integration of a dedicated *Review Agent* proved critical, consistently enhancing the robustness and accuracy of the final output through programmatic validation and self-correction mechanisms. Our evaluation, conducted on both the established non-spatial KaggleDBQA benchmark and a new, purpose-built spatial benchmark (SpatialQueryQA) featuring diverse geometries and complexities, demonstrated the framework's efficacy. The results confirm that our approach not only achieves high accuracy but, in several instances, generates queries that are more semantically aligned with user intent than those provided in the benchmarks themselves. While limitations persist, particularly in handling nuanced spatial operations like geodesic distance and complex aggregations, this research makes significant contributions to GIScience by effectively bridging the gap between intuitive natural language and the technical power of spatial SQL. It provides a generalizable framework for future autonomous GIS systems involving spatial databases and lays the groundwork for future research into interactive user clarification, advanced geometric reasoning, and the application of multi-agent architectures to other domain-specific SQL-based data retrieval challenges.

**Disclosure statement:** No potential conflict of interest was reported by the author(s).

**Data and Code Availability Statement:** All data used in this study, including benchmark questions, expected SQL queries, AI-generated queries (before and after Review Agent validation), and evaluation results are openly available on GitHub at: https://github.com/alikhosravi/Spatial-Text-to-SQL. The source code as well as a web-based user interface will be made available in a forthcoming update. The **Appendices** can be downloaded at https://bpb-us-e1.wpmucdn.com/sites.psu.edu/dist/0/172103/files/2026/03/Appendices.pdf